\documentclass[conference]{IEEEtran}
\usepackage[utf8]{inputenc}
\usepackage{amsmath}
\usepackage{amsfonts}
\usepackage{color}
\usepackage{natbib}
\usepackage[pdftex]{graphicx}
\usepackage{booktabs}       

\author{
  \IEEEauthorblockN{Rasmus Berg Palm}
  \IEEEauthorblockA{DTU Compute\\Tradeshift\\rapal@dtu.dk\\}  
  \and
  \IEEEauthorblockN{Florian Laws}
  \IEEEauthorblockA{Tradeshift\\fla@tradeshift.com}
  \and
  \IEEEauthorblockN{Ole Winther}
  \IEEEauthorblockA{DTU Compute\\olwi@dtu.dk}
}

\title{Attend, Copy, Parse \\
End-to-end information extraction from documents}

\begin{document}
\maketitle
\begin{abstract}

Document information extraction tasks performed by humans create
data consisting of a PDF or document image input, and extracted string outputs. This end-to-end data is naturally consumed and produced when performing the task because it is valuable \emph{in and of itself}. It is naturally available, at no additional cost.
Unfortunately, state-of-the-art word classification methods for information extraction cannot use this data, instead requiring word-level labels which are expensive to create and consequently not available for many real life tasks.
In this paper we propose the Attend, Copy, Parse architecture, a deep neural network model that can be trained directly on end-to-end data, bypassing the need for word-level labels.
We evaluate the proposed architecture on a large diverse set of invoices, and outperform a state-of-the-art production system based on word classification.
We believe our proposed architecture can be used on many real life information extraction tasks where word classification cannot be used due to a lack of the required word-level labels.\footnote{Code is available at \bf{github.com/rasmusbergpalm/attend-copy-parse}}
\end{abstract}

\section{Introduction}
As long as people communicate using unstructured documents, there'll be a demand for extracting structured information from these documents. However, extracting information from such documents is a tedious and costly task for humans. The field of information extraction investigates how to automate this task.

Consider employees at an enterprise processing invoices. They receive a paper or PDF invoice, extract a few important fields, e.g. the invoice number, total, due date, etc, and type it into a computer system. Simply doing their job they are implicitly creating a dataset consisting of pairs of PDF or paper invoices and their extracted fields. We define end-to-end data as such data that is naturally consumed and produced in a human information extraction tasks. By definition this data is available, should one wish to capture it.

Unfortunately such end-to-end data cannot be used with state-of-the-art machine learning methods for information extraction. Current state-of-the-art approaches require labeling of every word, which is costly to obtain, and consequently not available for many real life tasks. The distinction between end-to-end data and data labeled on the word level is subtle but important. In the invoice example the end-to-end data simply tells us \emph{what} the total is, whereas data labeled on the word level tells us \emph{where} it is. The former type of data is plentiful, produced naturally and is hard to learn from. The latter is scarce, must be explicitly produced for the purpose of machine learning, and is easier to learn from. 

In this paper we propose an end-to-end deep neural network architecture that can be trained directly from end-to-end information extraction data. This is our main contribution. We believe this architecture can be useful for many real-life information extraction tasks, where end-to-end data already exists, and word-level labeling is not feasible.

We evaluate our proposed architecture on a large diverse set of invoices. Invoices are somewhat special documents, in that documents from the same supplier often has a consistent layout, or template. Powerful methods exists for extracting information from templates, given that the template is known beforehand, but these methods generalize poorly across templates. Our proposed architecture addresses the harder task of learning a model that generalizes across document templates and as such can be used on unseen templates. This is important for invoices where the template often varies considerable across suppliers. We make the key assumption that the same structured information must be extracted from every document. For invoices this is a reasonable assumption, as invoices are fairly well defined documents.

Invoices are complex documents with considerable spatial structure, featuring both text and image modalities. The proposed architecture takes the spatial structure into account by using convolutional operations on the concatenated document text and image modalities. The text modality is represented in a principled manner by embedding the extracted text in a spatial grid. We assume the text of the document is given or extracted using an Optical Character Recognition (OCR) engine.

While this paper consider the case of invoices, the Attend, Copy, Parse framework is in no way limited to invoices. It could be used for any documents from which you are interested in extracting a fixed set of fields e.g. quarterly earning reports or meeting schedules from emails. 

\section{Related work}
\subsection{Pattern matching.} 
An intuitive approach to information extraction is to identify patterns in the unstructured data and use that to extract information. For instance, the total amount due in an invoice is typically to the right of a word that says ``total'', and is typically a decimal number, a pattern which can be captured using a regular expression.

There's a rich literature that expand upon this general idea. For instance, \citet{riloff1993automatically} suggests an expressive pattern matching languages that can take the syntactic sentence structure into account, e.g. match the noun of a given verb keyword and \citet{huffman1995learning} proposes extracting multiple target values using a single joint syntactic and keyword based pattern match. See \cite{muslea1999extraction} for a survey. 

For the more specific task of extracting information from business documents several works use a pattern matching approach. \citet{schuster2013intellix,rusinol2013field} and \citet{cesarini2003analysis} require users to annotate which words should be extracted for a given document template, then automatically generate patterns matching those words. At test time, these patterns generate candidate words, which are scored using heuristics. \citet{dengel2002smartfix,esser2012automatic} and \citet{medvet2011probabilistic} all use manually configured patterns based on keywords, parsing rules and positions. 

Pattern matching generally works better the more homogeneous and structured the input is. The main disadvantages are that the patterns takes time and expertise to create and maintain, and often doesn't generalize across document templates.

\subsection{Word classification.} 
Machine learning offers an elegant solution to deciding which words to extract. Given a dataset of documents and labels for each word, it becomes a regular classification task; given a word classify whether it should be extracted. If multiple values are to be extracted, e.g. total, date, etc. it becomes a multiclass classification task. The field of Natural Language Processing (NLP) uses this approach extensively, to perform a variety of tasks, e.g. Named Entity Recognition (NER) or part of speech tagging. See \citet{collobert2008unified} for an overview of tasks and methods. 

Traditionally the machine learning practitioner would come up with a set of features for words and use a shallow classifier, e.g. logistic regression, SVMs, etc. Many of the insights and heuristics used in the pattern matching approach can be re-purposed as features. However, state-of-the-art deep learning methods generally avoid feature engineering and favor word embeddings \citep{mikolov2013distributed, pennington2014glove} and deep neural networks \citep{ma2016end,lample2016neural,santos2015boosting}.

The main drawback of the word classification approach to information extraction is the need for a richly labeled dataset as every word must be labeled. Manual labeling is an expensive process, that is consequently not feasible for many real life information extraction tasks. 

Distant supervision \citep{mintz2009distant} proposes to generate the word-level labels heuristically from the available data. For instance, in our invoice example, if we know the total is \texttt{"200.00"} we can search for this string in the PDF, and label all words that match as the total. The CloudScan system proposed by  \citet{palm2017cloudscan} takes this approach and achieves state-of-the-art results on a large diverse set of invoices. CloudScan is the most comparable system, but it still requires word-level labels, they’re just generated using a distant supervision heuristic. In contrast the proposed method can train directly on the available end-to-end data. In general, the drawback of the distant supervision approach is that the quality of the labels depend entirely on the quality of the manually created heuristics. The heuristics should be smart enough to know that the 200 in the string \texttt{"total: 200 \$"} is probably the total, whereas 200 in the string \texttt{"200 liters"} is probably not. This is further complicated if the target string is not present letter-to-letter in the inputs.

\subsection{End-to-end methods}

Deep neural networks and the end-to-end training paradigm have lead to breakthroughs in several domains e.g. image recognition \citep{krizhevsky2012imagenet}, and machine translation \citep{bahdanau2014neural}. We are broadly inspired by these successes to investigate end-to-end learning for information extraction from documents.

Convolutional neural networks have been used extensively on document images, e.g. segmenting documents \citep{yang2017learning,chen2017convolutional,wick2018fully}, spotting handwritten words \citep{sudholt2016phocnet}, classifying documents \citep{kang2014convolutional} and more broadly detecting text in natural scenes \citep{liao2017textboxes,borisyuk2018rosetta}. In contrast to our task, these are trained on explicitly labeled datasets with information on \emph{where} the targets are, e.g. pixel level labels, bounding boxes, etc.

\citet{yang2017learning} proposes to combine the document image modality with a text embedding modality in a convolutional network for image segmentation, by fusing the modalities late in the network. We fuse the modalities as early as possible, which we find work well for our application.

The idea of reading from an an external memory using an attention mechanism was introduced in \cite{graves2014neural} and \citet{weston2014memory,sukhbaatar2015end}. Our memory implementation largely follows this paradigm, although it is read-only. External memories has since been studied extensively \citep{miller2016key,santoro2016one,graves2016hybrid}.

\section{Methods}
\begin{figure*}[!ht]	
    \centering \includegraphics[width=1.0\textwidth]{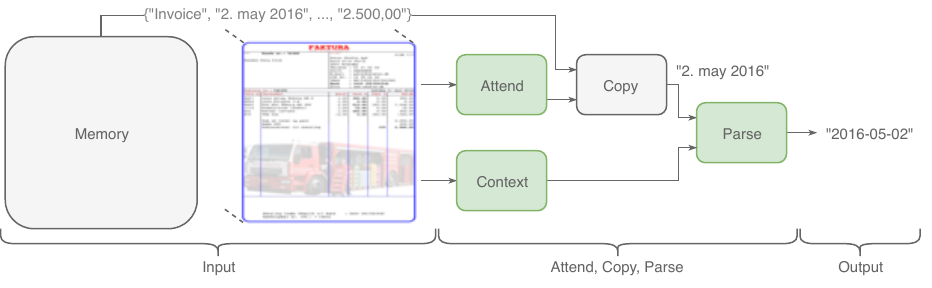}
    \caption{Overview of the Attend, Copy, Parse architecture. All modules are end-to-end differentiable. The modules highlighted in green are learned. The document image is $128 \times 128$ pixels, the same resolution the network sees. The dashed lines indicates additional feature channels, primarily text embeddings. See equation \ref{eq:features} for a list of all the features.}
    \label{fig:overview}
\end{figure*}

We are given a dataset of $N$ samples, each consisting of
\begin{itemize}
\item A document image $x \in [0,1]^{H \times W \times 3}$, where $H$ and $W$ is the height and width of the document image respectively.
\item A set of $P$ words $\mathbf{w} = \{(w_1, p_1), ..., (w_P, p_P)\} $ where $w_i$ is the word text and $p_i \in [0,1]^4$ denotes the normalized word position and size in the document image. This would typically be the output of an OCR engine applied to the document image.
\item Target strings $t_k$ for $K$ values we wish to extract, e.g. ``2018-07-23'' for a date.
\end{itemize}

The task is to learn a system that extracts the correct $K$ output strings $\mathbf{y} = [y_1,..., y_K]$ for a new input $(x, \mathbf{w})$.  The system should be able to handle:

\begin{itemize}
\item Unseen document templates, i.e. generalize across document templates.
\item Normalized target strings. Some target strings $t_k$, e.g. date and amounts, are normalized to standard formats in the end-to-end data, so may not be present letter-to-letter in the inputs.
\item Target strings spanning multiple words in the input. A single normalized target string, e.g. the date, \texttt{"2018-07-23"} might be represented as several words in the input e.g. \texttt{"23rd" "July," "2018"}.
\item Optional outputs. An output might be optional.
\end{itemize}

A natural human approach to information extraction is: for each value to extract: 1) locate the string to extract 2) parse it into the desired format. Our proposed framework is broadly inspired by this approach. On a high level the Attend, Copy, Parse framework produces an output string $y_k$ as
\begin{align*}
	a_k &= \text{Attend}_k\left(x, \bf{w}\right) \\
    c_k &= \text{Copy}_k\left(a_k,\bf{w}\right) \\
    h_k &= \text{Context}_k\left(x, \bf{w}\right) \\
   	y_k &= \text{Parse}_k\left(c_k, h_k\right) \ .       
\end{align*}
The main idea is to build an external memory bank the same size as the document image, containing the words encoded as a sequence of characters at the memory positions corresponding to the positions of the words in the image. The attend module attends to this memory bank and the copy module copies out the attended string. The parse module parses the attended string into the desired output format, optionally given a context vector computed from the inputs. See figure \ref{fig:overview} for an overview.

We will now describe each module in detail for a single model. The $k$ subscript is dropped in the following since we train separate models for each of the $K$ fields.

\subsection{External memory}
We start by constructing the external memory bank, $M \in \{0,1\}^{H \times W \times G \times L \times D}$ containing N-grams up to length $G$. In our experiments $G=4$. The N-grams are created by an algorithm that divides the document into a number of non-intersecting lines and sorts the words by their horizontal position inside each line. 

The N-grams are encoded as a sequence of one-hot encoded characters, such that $L$ is the maximum sequence length we consider and $D$ is the size of the character set we're using. For our experiments $L=128$ and $D=103$. The memory has $W \times H \times G$ slots, that can each contain an encoded N-gram. The first two dimensions correspond to the spatial dimensions of the document and the third to the length of the N-gram. This memory tensor quickly becomes too big to keep in memory. However, since it is very sparse it can be represented as a sparse tensor.

It is not immediately obvious which slots in the memory should contain the N-grams, since each N-gram span multiple pixels in the document image. We found that it suffices to store the encoded N-grams in the single slot corresponding to the top-left corner position of the first word in each N-gram. This makes the sums in the copy module considerably faster.

\subsection{Attend}

We compute unnormalized attention logits $u \in \mathbb{R}^{H \times W \times G}$ for each slot in the external memory.
\begin{align}
 u &= \text{Attend}\left(x, \bf{w}\right) \ .
\end{align}
Since we know which slots in the memory are not empty, we only want the network to put probability mass here. To achieve this we set $u$ to $-1000$ everywhere else before we compute the attention distribution $a \in [0,1]^{H \times W \times G}$ using the softmax operation.
\begin{align}
\label{eq:attend}
a_{ijg} &= \frac{e^{u_{ijg}}}{\sum_{i=1}^{H} \sum_{j=1}^{W} \sum_{g=1}^{G} e^{u_{ijg}}} \ .
\end{align}
In our experiments we parameterize the $\text{Attend}$ function in the following way. We construct an input representation of the document $r \in \mathbb{R}^{H \times W \times U}$, where $U$ is the number of feature channels. 
\begin{align}
	\label{eq:features}
	r &= \text{Concat}\left(x, q_w, q_p, q_c, z, \delta_x, \delta_y, \eta \right) \ ,
\end{align}
where $q_w$, $q_p$ and $q_c$ are learned 32 dimensional word, pattern and character embeddings, respectively. The pattern embedding is an embedding of the word after all characters have been replaced with the character \texttt{x}, all digits with \texttt{0} and all other characters with a dot. Word and pattern embeddings are replicated across the spatial extent of the entire word. Character embeddings are replicated across the spatial extent of each character in the word. In case characters overlap due to downsampling of the document image, the rightmost character wins. $z$ is two binary indicator channels whether the N-gram at this position parses as an amount or a date, according to two pre-configured parsers. $\delta_x$ and $\delta_y$ contain the normalized ([0,1]) horizontal and vertical positions. Finally $\eta$ is a binary indicator whether the external memory at this spatial position is non-empty. 

\begin{figure}[!ht]	
    \centering \includegraphics[width=0.4\textwidth]{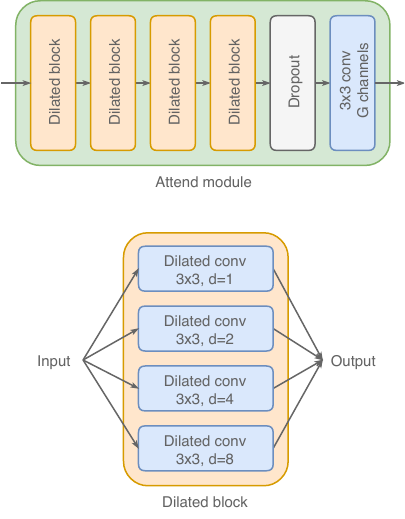}
    \caption{Attend module and dilated block details. ``d'' denotes the dilation rate.}
    \label{fig:dilated}
\end{figure}

We pass $r$ through four dilated convolution blocks \citep{yang2017learning}. A dilated convolution block consists of a number of dilated convolution operations, each with a different dilation rate, that all operate in parallel on the input, and whose outputs are concatenated channel wise. Each dilated convolution block contains 4 dilated convolution operations with dilation rates $[1,2,4,8]$, each with 32 $3 \times 3$ filters with ReLU nonlinearities. The output of each dilated convolution block has 128 channels. See figure \ref{fig:dilated}.

The dilated convolution block allows the network to preserve detailed local information through the layers with smaller dilation rates, while capturing a large context through the layers with higher dilation rates, both of which are important for our task.

After the 4 dilated convolution blocks we apply dropout, and a final convolution operation with $G$ linear $3 \times 3$ filters, to get the unnormalized attention logits $u$, for each memory slot. We tried several other attention module architectures including U-nets \citep{ronneberger2015u}, residual networks \citep{he2016deep} and deeper, wider variants of the residual blocks. The architecture described was chosen based on its performance on a validation set.

\subsection{Copy}
Given the memory $M$ and the attention distribution we copy out the attended N-gram $c \in [0,1]^{L \times D}$,
\begin{align}
\label{eq:copy}
	c_{ld} &= \sum_{i=1}^{H} \sum_{j=1}^{W} \sum_{g=1}^{G}  a_{ijg} M_{ijgld} \ .
\end{align}
We use the term copy, although it is a soft differentiable approximation to copy, such that each $c$ is a weighted sum over all the N-grams present in the document. Accordingly, each character $c_{l}$ is a distribution over all the characters in the character set.

\subsection{Context}
The context of an N-gram is often important for parsing it correctly. For instance the date ``02/03/2018'' is ambiguous. It can either be the 2nd of March or the 3rd of February, but the context can disambiguate it. For instance, if the language of the document is German, then it is more likely that it is the former. In our experiments we use the following context function. 
\begin{align*}
	h_f &= \sum_{i=1}^{H} \sum_{j=1}^{W} \sum_{g=1}^{G} a_{ijg} v_{ijf} \ ,
\end{align*}
where $v \in \mathbb{R}^{H \times W \times 128}$ is the output of the last dilated convolution block of the attend module, after dropout. Thus, $h$ is a vector $h \in \mathbb{R}^{128}$. For simplicity the implementation of the context function in our experiments only depend on the attention module, but in general, it can be any function of the input $(x, \bf{w})$.

\subsection{Parse}
Given the attended word $c$, and the context vector $h$, we parse $c$ into the output $y$. This is in essence a character based sequence to sequence task:
\begin{align}
	y = \text{Parse}(c, h) \ .
\end{align}
The implementation of $\text{Parse}$ depends on the output format. Some fields might not need any parsing, while dates and amounts need different parsing. In the following we describe the four different parsers we use.

\textbf{NoOp Parser.} This is the simplest parser. It returns the attended sequence as is
\begin{align*}
	y = c \ .
\end{align*}
\textbf{Optional Parser.} This returns a mixture of the attended input and a string $\epsilon$ consisting solely of \texttt{<EOS>} tokens
\begin{align*}
	\alpha &= f(h) \\
	y &= (1-\alpha) c + \alpha \epsilon \ ,
\end{align*}
where $\alpha$ in $[0,1]$ is the mixture parameter. We use a single fully connected layer with one sigmoid output unit for $f$.

\textbf{Date Parser.} Our target dates are represented in ISO 8601 format, e.g. ``2018-07-23'' which has a constant output length of ten characters, with two fixed characters. We use the following architecture in our experiments which we find work well:
\begin{align*}
	e &= \sum_{l=1}^{L^*} \text{CNNMP}(c) \\
	y &= \text{MLP}\left(\text{Concat}(e,h)\right) \ ,
\end{align*}
where $\text{CNNMP}$ is 4 layers of CNN with 128 ReLU units with kernel size 3, followed by stride 2 maxpooling, $L^* = \frac{L}{2^4}$ is the length of the sequence after the four maxpooling layers. As such $e \in \mathbb{R}^{128}$. MLP is 3 fully connected layers with 128 ReLU units, followed by dropout, and finally a linear layer with $10 \times D$ linear outputs, which outputs the unnormalized logits for the ten characters. We could use a smaller output character set than $D$, but for simplicity we use a single character set for representing all characters.

\textbf{Amount Parser.} The amounts in the dataset are normalized by removing leading and trailing zeros, e.g. ``00.020'' gets formatted to ``0.02'' and ``1.00'' to ``1''. We use a fixed output size of $16$ characters.

Our amount parser is a pointer-generator networks \citep{see2017get}, with a bidirectional LSTM with 128 units to encode the input sequence, and a LSTM with 128 units to generate the output hidden states. The idea behind using a pointer-generator network is that if the attended input is a digit, it can be copied directly, if it is a decimal separator, the network can generate a dot, and if it is the last significant digit the network can generate an \texttt{<EOS>} token. For details see the appendix.

\subsection{Loss and attention regularization}
The main loss is the average cross-entropy between the targets characters $t$ and the output characters $y$.

We found that the softmax operation in equation (\ref{eq:attend}) had a tendency to underflow for many of the non-empty memory positions causing the network to get stuck in the initial phase of training. To solve this we added a regularization term to the loss defined as the cross-entropy between a uniform attention distribution over the non-empty memory positions and the attention distribution produced by the attend module in equation (\ref{eq:attend}):
\begin{align*}
	\mathcal{L}(y, t) = -\frac{1}{L}\sum_{l=1}^L \log(y_l[t_l]) - \lambda \frac{1}{|I|}\sum_{(i,j,g) \in I} \log(a_{ijg}) \ , 
\end{align*}
where $y_l[t_l]$ indicates the $t_l$th element of $y_l$, $I$ is the set of non-empty memory positions and $\lambda$ is a small scalar hyper-parameter. The regularization encourages the network to place attention mass uniformly on the non-empty memory positions.

\section{Experiments}
Our dataset consists of 1,181,990 invoices from 43,023 different suppliers. Approximately 37\% of the invoices are scanned document images, the rest are digital PDF files. The dataset is obtained from production usage of an invoice information extraction system where the suppliers upload a document and the system extract several fields of interest. If the system makes a mistake the suppliers type in the correct value for the field. Note, the suppliers do not need to indicate which word in the document corresponds to the field so as to not burden them with additional labeling effort. As such the dataset fits our end-to-end dataset definition and the description in the methods section; it is not known which words in the document corresponds to the fields we wish to extract, we simply know the target value of each field. We focus on seven important fields, 1) the invoice number, 2) the order number, 3) the date, 4) the total, 5) the sub total before tax, 6) the tax total and 7) the tax percent.

We split the suppliers randomly into 42,163 training suppliers and 860 testing suppliers. The training and testing sets of documents consists of all documents from each set of suppliers, 1,153,078 and 28,912 documents, respectively. Splitting on the suppliers instead of splitting on the documents allows us to measure how well the model generalize across document templates, assuming all suppliers use different templates.

There are two main sources of noise in this dataset, that will limit the maximum performance of any system.
\begin{enumerate}
\item OCR. If the OCR engine makes a mistake, it is very hard to recover from that later on. With a powerful enough parser it is not impossible, but still unlikely.
\item Suppliers. The suppliers can and will type in field values that are not present in the document image. For instance the supplier might type in the date they are correcting the document, rather than the date present in the document, in order to not backdate the invoice. This kind of noise is near impossible to correct for.
\end{enumerate}

In order to estimate the maximum performance of a system for this dataset, we perform the following analysis. Given a target value $t_k$ we try to find this value in the document. If $t_k$ is an amount, we use a recall oriented amount parser based on regular expressions to parse the strings in the document. We apply a similar procedure for dates. This way we can measure an approximate fraction of target strings present in the document. We use the term ``Readable'' for this fraction of target values. For the fields that do not use a parser, i.e, the invoice number and the order number, this is an upper bound on the possible accuracy for our proposed end-to-end architecture. For the other fields, it is conceivable that we can learn a better parser than the regular expression based parsers, but it is still a decent approximation of the achievable accuracy.

We compare the proposed end-to-end system to the production system described in \citet{palm2017cloudscan}, which is trained and tested on the same data. In short the production system uses the word-classification approach with word labels derived using a distant supervision heuristic. A logistic regression classifier classifies each word into, e.g. the invoice number, total, etc. After classifying the words, they are parsed, and the system uses heuristics to pick the best candidate words, e.g. for the total fields, it tries to find a joint solution such that the totals add up, etc.

We train a separate model for each of the seven fields, which all have the same architecture and hyper-parameters, except for the parsers. The invoice number uses the no-op parser, the order number the optional parser, the date the date parser, and the rest use the amount parser. We pre-train the amount parser on amounts extracted from the training documents, parsed with a conventional regular expression based parser. We observe that the amount parser quickly learns to replicate the regular expressions. Each model is optimized using the Adam \citep{kingma2014adam} optimizer, with a batch size of 32, learning rate $0.0003$, dropout of 0.5, $\lambda = 0.0001$, and trained for 50,000 batch updates, with a small L2 regularization of $0.0001$. We resize the document image to $128 \times 128$ pixels, disregarding aspect ratio. See table \ref{table:results} for the results.

\begin{table}[!ht]
\caption{Results. Fraction of correct values. 
}
\label{table:results}
\centering
  \begin{tabular}{lcccc}
  \toprule
  Field & Readable & Prod & Prod- & Attend, Copy, Parse \\  
  \midrule
    Number & 0.90 & 0.78 & 0.78 & \bf{0.87}\\
    Order id & 0.90 & 0.82 &  0.82 & \bf{0.84}\\
    Date & 0.83 & 0.70 & 0.70 & \bf{0.80}\\
    Total & 0.81 & \bf{0.85} & 0.77 & 0.81\\
    Sub total & 0.84 & \bf{0.84} & 0.73 & 0.79\\
    Tax total & 0.80 & \bf{0.87} & 0.77 & 0.80\\
    Tax percent & 0.79 & 0.83 & 0.68 & \bf{0.87} \\
  \midrule
    Average & 0.84 & 0.81 & 0.75 & \bf{0.83} \\
  \bottomrule
  \end{tabular}  
\end{table}

The proposed Attend, Copy, Parse system performs better on average, and close to the approximate maximum accuracy possible given the architecture, denoted ``Readable''. The invoice number is a good field to compare how good the respective systems are purely at ``finding'' the correct N-gram, since there's no parsing involved in either system. Here the Attend, Copy, Parse system excels. However, the production system, ``Prod'', performs significantly better on 3 of the 4 amount fields. The production system uses a heuristic to pick the total fields jointly, such that they add up. In order to test how much this heuristic improves the results we test a version of the production system with this heuristic disabled. We denote this ``Prod-''. This version simply choose the word with the highest probability given by the logistic regression classifier that can be parsed into an amount for each of the total fields. It is clear from the results that this heuristic considerably boosts the accuracy on the total fields. Interestingly the Attend, Copy, Parse architecture recovers more correct tax percentages than can be found in the documents. Upon closer inspection this is because many documents have zero taxes, but do not contain an explicit zero. The pointer-generator amount parser learns to generate a zero in these cases.

\section{Discussion}

We have presented a deep neural network architecture for end-to-end information extraction from documents. The architecture does not need expensive word-level labels, instead it can directly use the end-to-end data that is naturally produced in a human information extraction tasks. We evaluated the proposed architecture on a large diverse set of invoices, where we outperform a state-of-the-art production system based on distant supervision, word classification and heuristics.

The proposed architecture can be improved in several ways. The most obvious shortcoming is that it can only handle single page documents. This is theoretically easy to remedy by adding a new page dimension to the inputs, turning the spatial convolutions into volumetric convolutions, and letting the attend module output attention over the pages as well. The main concern is the computational resources required for this change.

It should be possible to learn a single network which output the $K$ strings. We experimented with this, by letting the attention module output $K$ attention distributions, having $K$ separate copy and parse modules, and training everything jointly using the sum of losses across each of the $K$ outputs. It worked, but less well. We suspect it is because of imbalance between the losses. For instance dates have lower entropy in general compared to invoice numbers. Loss imbalance is a general challenge in the multi-task learning setting \citep{kendall2017multi, chen2017gradnorm}.

Taking the dependencies between the total fields into account (Prod) significantly increases the performance of the system even given a relatively weak classifier. Unfortunately, the heuristic used in the Prod system cannot be directly used with the Attend, Copy, Parse architecture as it is not differentiable.

We did come up with an idea to incorporate the two constraints (total = sub total + tax total and tax total = sub total $\cdot$ tax percentage) in our Attend, Copy, Parse framework but it did not improve on the results. Here we describe the idea briefly. It consists of three steps: 1) let the network output a probability that each total field should be inferred from the constraints instead of being outputted directly, 2) Assuming you will sample which fields to infer from this distribution, write up the marginal probability of all the fields being correct and 3) Use the negative log of this probability as a loss instead of the four individual total field losses. If you sample three or four fields to infer, then the probability of all the fields being correct is zero, since you can at most infer two of the fields from the constraints. If you sample two fields or less, then the probability that all the fields are correct is the probability that all the non-inferred fields are correct. The marginal probability that all of the fields are correct is then the sum over the probability that a permutation of fields to be inferred is chosen, multiplied by the probability that all the non-inferred fields for the given permutation are correct. There's only $\binom{4}{0} + \binom{4}{1} + \binom{4}{2} = 11$ permutations that give non-zero probabilities, so they can simply be computed and summed.

The presented architecture can only extract non-recurring fields as opposed to recurring, structured fields such as invoice lines. Theoretically it should be possible to output the lines by recurrently outputting the fields of a single line at a time, and then conditioning the attention module on the previously outputted line. This would be an interesting direction for future work.

\bibliography{refs}
\bibliographystyle{plainnat}

\section{Appendix}
\subsection{Amount parser details}
We use a fixed output size of $O=16$ characters. Each output character $y_o \in [0,1]^D$ is a weighted sum over characters from the attended string $c$, or generated from the character set.
\begin{align*}
	y_o &= \rho_o \sum_{l=1}^L (a_{ol} c_{l}) + (1 - \rho_o) g(e_o) \ ,
\end{align*}
where $\rho_o \in [0,1]$ determines whether $y_o$ should be copied from the attended string $c$ or generated from the character set. $a \in [0,1]^{O \times L}$ is $O$ attention distributions over each character in the input string; one for each character in the output string, which determines which character to copy. $g$ is a function that maps $e_o \in \mathbb{R}^{256}$ to a distribution over the character set. In our experiments we use a single dense layer with $D$ outputs and a softmax nonlinearity. $\rho_o$ is given as
\begin{align*}
	\rho_o &= f(e_o) \ ,
\end{align*}
where $f$ is a function that maps $e_o$ to $[0,1]$. In our experiments we used a single dense layer with a single sigmoid output. $e_o$ is
\begin{align*}
	e_o &= \sum_{l=1}^L (a_{ol} h_{l}) \ ,
\end{align*}
where $h \in \mathbb{R}^{L \times 256}$ is the encoded input string $c$. We use a bidirectional LSTM with 128 hidden units each to encode the input string,
\begin{align*}
	h &= \text{BiLSTM}(c_{k}) \ .
\end{align*}
All that remains to be defined are $a$, the $O$ attention distributions over the input characters. 
\begin{align*}
	a_{ol} &= \frac{e^{\alpha_{ol}}}{\sum_{l=1}^L e^{\alpha_{ol}}} \\
	\alpha_{ol} &= \phi(u_o, h_l) \ ,   
\end{align*}
where $\phi$ maps $u_o \in \mathbb{R}^{128}$ and $h_l \in \mathbb{R}^{256}$ to $R$. Following \citet{bahdanau2014neural} we use
\begin{align*}
	\phi(u_o, h_l) &= \text{tanh}(u_o W_u + h_l W_h) W_t \ ,
\end{align*}
where $\text{tanh}$ is applied element-wise and $W_u \in \mathbb{R}^{128 \times 128}$, $W_h \in \mathbb{R}^{256 \times 128}$ and $W_t \in \mathbb{R}^{128 \times 1}$ are learned weight matrices. Finally, $u \in \mathbb{R}^{O \times 128}$ is the output of a 128 unit LSTM run for O steps with no inputs.

\end{document}